\documentclass[Afour,sagev,times,doublespace]{sagej}

\usepackage{multirow}
\usepackage{graphicx}
\usepackage{caption}
\usepackage{subcaption}  % \usepackage{subfig}
\usepackage{amsmath}
\usepackage{moreverb,url}
\usepackage[colorlinks,bookmarksopen,bookmarksnumbered,citecolor=red,urlcolor=red]{hyperref}
\usepackage{ulem} % for \sout
\usepackage{xcolor} % to set color of content

\newcommand\BibTeX{{\rmfamily B\kern-.05em \textsc{i\kern-.025em b}\kern-.08em
T\kern-.1667em\lower.7ex\hbox{E}\kern-.125emX}}

\def\volumeyear{2016}

\begin{document}

%\runninghead{Smith and Wittkopf}

\title{Convolutional Neural Networks for Accurate Measurement of Train Speed}

\author{Haitao Tian\affilnum{1}, Argyrios Zolotas\affilnum{1} and Miguel Arana-Catania\affilnum{2,1}}

\affiliation{\affilnum{1} Faculty of Engineering and Applied Sciences, Cranfield University\\
\affilnum{2} Digital Scholarship at Oxford, University of Oxford}

\corrauth{Miguel Arana-Catania}

\email{humd0244@ox.ac.uk}

% Abstract and Keywords
\begin{abstract}

In this study, we explore the use of Convolutional Neural Networks for improving train speed estimation accuracy, addressing the complex challenges of modern railway systems. We investigate three CNN architectures — single-branch 2D, single-branch 1D, and multiple-branch models — and compare them with the Adaptive Kalman Filter. We analyse their performance using simulated train operation datasets with and without Wheel Slide Protection activation. Our results reveal that CNN-based approaches, especially the multiple-branch model, demonstrate superior accuracy and robustness compared to traditional methods, particularly under challenging operational conditions. These findings highlight the potential of deep learning techniques to enhance railway safety and operational efficiency by more effectively capturing intricate patterns in complex transportation datasets.

\end{abstract}


\maketitle

%% Main Matter

\section{Introduction}
\label{chap:Introduction}

% \subsection{Research background}
% Accurately estimating train speed is a critical aspect of modern railway systems, with far-reaching implications for safety, efficiency, and overall performance. The importance of precise speed measurement cannot be overstated, as it underpins numerous vital functions in train operation. At the forefront is safety, where accurate speed data is essential for the Automatic Train Protection (ATP) system to calculate safe braking distances, enforce speed limits, prevent collisions, and activate emergency braking when necessary. Even minor errors in speed estimation could lead to catastrophic accidents, especially at high speeds \citep{wang2018SafetyTheoryControl}.

Accurately estimating train speed is a critical aspect of modern railway systems, with far-reaching implications for safety, efficiency, and overall performance. Accurate speed data is essential for the Automatic Train Protection (ATP) system to calculate safe braking distances, enforce speed limits, prevent collisions, and activate emergency braking when necessary. Even minor errors in speed estimation could lead to catastrophic accidents, especially at high speeds \citep{wang2018SafetyTheoryControl}.

Beyond safety, accurate speed control plays a crucial role in energy efficiency, allowing trains to operate at optimal speeds and minimise unnecessary acceleration and braking \citep{peng2024EnergyefficientTrainControl,davoodi2023EnergyManagementSystems}. This not only reduces energy consumption but also contributes to environmental sustainability in the transportation sector. Precise speed information is equally vital for maintaining schedules and ensuring punctuality, facilitating better coordination between trains, and optimising overall network efficiency \citep{wang2021IntegratedEnergyefficientTrain}. Passenger comfort is another beneficiary of accurate speed control, as smooth acceleration and deceleration contribute to a pleasant riding experience \citep{peng2022ReviewPassengerRide}. Furthermore, infrastructure management is highly dependent on accurate speed data to plan maintenance activities and assess track conditions, as wear patterns are closely linked to train speeds \citep{sedghi2021TaxonomyRailwayTrack}. Lastly, railways must adhere to strict speed limits and regulations, making accurate speed measurement necessary for demonstrating compliance and avoiding penalties \citep{lu2019DiscussionTechnologiesImproving}.

Despite its importance, train speed estimation faces numerous challenges that complicate the quest for accuracy. The traditional method of measuring train speed through wheel revolutions is particularly problematic due to wheel wear and rolling contact fatigue \citep{ekberg2014WheelRailRolling,shi2018FieldMeasurementsEvolution}. As wheels decrease in diameter over time due to wear or reprofiling, speed calculations become increasingly inaccurate without frequent recalibration. This issue necessitates regular maintenance stops, introduces gradual drift in accuracy between calibrations, and can be affected by variations in wear between different wheels. Environmental factors pose another significant challenge, as train speed sensors are ``susceptible to diverse operating conditions such as snow, rain, fog, tunnel, hilly region, slip, slide, etc." \citep{muniandi2019TrainDistanceSpeed}. These conditions can interfere with various sensing technologies, impacting the reliability of speed measurements across different terrains and weather conditions.

The phenomena of wheel slip and slide during acceleration or braking, particularly in adverse weather, can lead to substantial errors in wheel-based speed measurements \citep{moaveni2020SupervisoryPredictiveControl,namoano2022DataDrivenWheelSlip}. This challenge is compounded by the complexity of integrating data from multiple sensors, each with its own sampling rates and noise characteristics \citep{zhan2021ResearchSpeedSensor}. Real-time processing requirements add another layer of difficulty, as high-speed trains demand rapid, instantaneous speed estimation for safe operation \citep{zhou2021RealtimeEnergyefficientOptimal}. This necessitates efficient algorithms and processing capabilities to handle data from multiple sensors with minimal latency. Cost constraints also play a role, as highly accurate speed measurement systems may be prohibitively expensive for widespread deployment across entire rail networks. Moreover, any speed estimation system must be extraordinarily reliable and robust, capable of functioning consistently in the harsh railway environment, which includes vibrations, electromagnetic interference, and temperature variations \citep{yaghoubi2019HighSpeedRail}.

% The development of more accurate and reliable non-contact speed measurement technologies could eliminate issues associated with wheel wear. Potential technologies in this area include advanced Doppler radar systems, optical speed sensors using computer vision, and magnetic field-based speed sensing\citep{du2020SpeedCalibrationTraceability,nagaraju2024RealtimeImplementationOptical,arif2014MeasurementSpeedCalibration}. Improved integration of GNSS with Inertial Navigation Systems (INS) and enhanced map-matching algorithms could further refine overall speed estimation accuracy, particularly when combined with other sensor data \citep{kim2015HighspeedTrainNavigation,liu2023RobustTrainLocalisation}. The concept of distributed sensing, utilising a network of trackside sensors to provide additional speed data, offers another avenue for improving accuracy and reliability \citep{wiesmeyr2020RealTimeTrainTracking}. Emerging technologies such as blockchain could be explored for secure logging of speed measurements, ensuring data integrity for regulatory compliance \citep{rusch2022ZugChainBlockchainBasedJuridical}. Edge computing solutions implemented on trains could enable more complex, real-time processing of sensor data, facilitating more sophisticated fusion and estimation algorithms \citep{zhao2021ContinuousMonitoringTrain}. Additionally, the development of self-calibrating systems that can automatically detect and compensate for wheel wear or other systematic errors could significantly reduce the need for manual recalibration \citep{zhang2024MountingMisalignmentTime}.

As railway systems continue to evolve, with trends towards higher speeds and increased automation, the importance of precise speed measurement will only grow. The integration of multiple sensor types, coupled with sophisticated algorithms and potentially machine learning (ML) approaches, promises to provide more robust and accurate speed estimates. Continued research and development in this area, leveraging advances in sensor technology, data processing, and artificial intelligence, will be crucial for ensuring the safety, efficiency, and performance of future rail transport systems. By addressing the challenges of accurate speed estimation, the railway industry can move towards more reliable, cost-effective, and technologically advanced solutions that support the ongoing evolution of train operations and safety systems.

A particularly relevant type of machine learning architecture is the Convolutional Neural Networks \citep{atlas1987artificial,denker1988neural,lecun1989backpropagation,lecun1998gradient}. These are a class of neural networks particularly well-suited for processing data with spatial structures. In convolutional layers, neurons are connected only to a local region of the input data, known as the receptive field, which helps capture local features of the input. This local processing of the input significantly reduces the number of parameters in the network and lowers the model's complexity as compared with a traditional fully-connected neural network. Each convolutional kernel acts as a feature extractor, learning appropriate weights through training to detect specific features in the input. Multiple convolutional kernels can operate simultaneously, generating multiple feature maps, with each map capturing different features from the original input, which are then combined to produce the network output.

In this article we develop this latter proposal, applying machine learning approaches to train speed estimation. Currently, accurately predicting train speed predominantly relies on traditional sensor fusion methods, and the application of machine learning in this area remains relatively limited. This work develops a robust and adaptable methodology designed to enhance the accuracy and reliability of speed measurements under various operational conditions, including training data from challenging environments with low adhesion. Another key contribution is the exploration and rigorous evaluation of different Convolutional Neural Network (CNN) architectures, leading to the identification of the optimal model for speed estimation. The comparative analysis between CNN-based models and the Adaptive Kalman Filter (AKF), highlights the potential for machine learning to outperform conventional approaches. Ultimately, this research provides insights that could significantly improve train safety, efficiency, and operational performance across diverse railway environments.

To summarise, the contributions of this paper to the estimation of train speeds include:

\begin{itemize}
    \item Developing a robust machine learning algorithm for accurate speed measurement under various conditions.
    \item Exploring and evaluating multiple CNN architectures.
    \item Demonstrating machine learning's potential to outperform traditional approaches like the Adaptive Kalman Filter.
\end{itemize}

% The paper is organised as follows. Section 1 introduces the challenges of speed estimation in the railway industry and sets the research objectives. Section 2 reviews existing speed prediction methods, analysing conventional and machine learning approaches. Section 3 details the data collection and pre-processing methodology. Section 4 explores and compares three CNN architectures, benchmarking against the AKF. The final section summarizes key findings and suggests future research directions.

\subsection{Related Work}

Multi-sensor fusion refers to the integration of data from multiple sensors to enhance the accuracy, reliability, and robustness of the system's performance. This approach is particularly beneficial in applications where single sensors may be susceptible to noise, errors, or limitations in measurement range. In the context of speed measurement, multi-sensor fusion leverages data from various sources, such as GPS, wheel speed sensors, and inertial measurement units (IMUs), to provide a more accurate and comprehensive estimation of the vehicle's speed. By combining these different data streams, the system can compensate for the weaknesses of individual sensors, such as the susceptibility of GPS to signal loss in urban canyons or the wheel speed sensor's sensitivity to skidding \citep{liu2024DeepGPSDeepLearning,zhaotingyang2024ResearchDetectionCorrection}. Techniques like Kalman filtering and particle filtering are commonly employed to fuse these sensor data effectively, producing a unified output that is more reliable than any single sensor's data. This methodology enhances the precision of speed estimation, making it invaluable for applications in autonomous driving, traffic management, and safety-critical systems where accurate speed information is essential.

\subsubsection{Sensor fusion for speed estimation.}
\label{sensor_fusion_for_speed_estimation}

Studies on speed estimation and vehicle localisation present a variety of methods and technologies, each with its unique contributions and drawbacks. A common thread across these studies is the utilisation of Kalman Filter techniques to enhance the accuracy of speed and position measurements, although the specific implementations and sensor integrations differ \citep{govaers2019IntroductionImplementationsKalman,cunillera2022RealtimeTrainMotion,pichlik2016TrainVelocityEstimation}. 

\cite{deng2020MultisensorBasedTrain} introduced the Covariance Intersection algorithm to integrate data from multiple sensors, including GPS, Inertial Navigation Systems (INS), and speed sensors. The AKF was used to adjust the measurement noise covariance matrix, thereby mitigating the effects of uncertain environmental factors. In the event of a GPS outage, the train's speed and position are estimated using the Velocity Sensor/INS scheme. Similarly, \cite{ernest2004TrainLocatorUsing} tackled the railway vehicle localisation problem by fusing data from odometers and accelerometers using a Kalman Filter. However, the long-term accuracy of his method was heavily dependent on the precision of the odometer and correct detection of wheel slip, illustrating a common challenge in relying on wheel-based sensors.

In contrast, \cite{tanelli2006LongitudinalVehicleSpeed} presented a state estimator algorithm that solely used data from hall-effect wheel encoders and a single-axis accelerometer. This method, notable for its low computational cost, could be implemented in commercial vehicle control units and was extensively tested under various driving conditions. \cite{mei2008MeasurementVehicleGround} adopted a different approach by using inertial sensors on the railway vehicle bogie frame to measure dynamic responses and extract wheelset motion features. This method avoided issues related to wheel slip or slide but showed significant errors at low speeds, indicating a trade-off between sensor type and measurement accuracy.

\cite{danhu2009ResearchInformationFusion} introduced an extended Kalman Filter algorithm that fused data from multiple sensors, including accelerometers, gyroscopes, and wheel speed sensors, to estimate vehicle speed and road adhesion characteristics. Their preliminary results were promising, though largely based on simulations, underscoring the need for extensive field testing. \cite{chu2011DesignLongitudinalVehicle} combined fuzzy logic with Kalman Filters to estimate vehicle longitudinal speed, achieving an error margin within 4\%. This hybrid approach demonstrated the potential benefits of integrating multiple estimation techniques.

\cite{malvezzi2014LocalizationAlgorithmRailway} used tachometers and inertial measurement units (IMU) to improve speed and distance estimation on the Italian railway network, particularly under critical adhesion conditions. Their Kalman Filter-based method showed significant improvements over the existing algorithm in the Italian network, though it struggled with slow initial acceleration. This study emphasised the importance of algorithmic refinement to handle varying operational conditions.

\cite{kim2015SlipSlideDetectiona} proposed a two-stage federated Kalman Filter for high-speed train navigation systems, incorporating multiple sensors and an adaptive information-sharing algorithm to mitigate severe performance degradation from sensor errors. This approach highlighted the benefits of federated filters in managing sensor discrepancies and enhancing system robustness. Finally, \cite{muniandi2019TrainDistanceSpeed} introduced a probabilistic weighted fusion algorithm based on a nonlinear longitudinal train dynamic model. This method combined state estimates from distributed and sensor-specific extended Kalman Filters, using measurements from a diverse set of sensors to improve distance and speed estimations.

While most studies leveraged the Kalman Filter or its variants to enhance speed estimation accuracy, their approaches varied significantly in sensor integration and application contexts. The common challenge of sensor dependency and error management underscores the need for continued innovation and hybrid methodologies to achieve robust and reliable speed measurements across diverse railway and vehicular systems.

\subsubsection{Machine learning for speed estimation.}
\label{sec:machine_learning_for_speed_estimation}

Recent advancements in speed estimation methods have seen a shift towards leveraging machine learning algorithms alongside conventional sensor fusion techniques. ML approaches, such as regression models, decision trees, random forests, neural networks, and deep learning models, are gaining popularity due to their ability to predict complex nonlinear relationships from extensive datasets without requiring explicit modelling of object dynamics \citep{jordan2015MachineLearningTrends, choi2020IntroductionMachineLearning, alpaydin2021MachineLearning}. Unlike traditional sensor fusion methods such as the Kalman Filter which rely on predefined models and assumptions about object behaviour, ML methods offer a more flexible framework. They can integrate data from various sensors installed on moving objects, making them suitable for environments with nonlinear and non-Gaussian noise characteristics \citep{blasch2021MachineLearningArtificial,brena2020ChoosingBestSensor}.

Supervised machine learning, a subset of ML, dominates speed estimation tasks where known speed serves as labelled data. Models are trained to map sensor inputs (e.g., accelerometer readings, positional data) to speed labels, enabling accurate speed predictions in diverse scenarios. This approach contrasts with unsupervised machine learning, which discovers patterns without labelled data, and reinforcement learning, which optimises behaviour through interaction with the environment. \cite{shafique2015UseAccelerationData} explored SVM, AdaBoost, decision trees, and random forests to classify transportation modes, such as walking, bicycle, train or car, using accelerometer data, emphasising the impact of preprocessing and training data generation on prediction accuracy.

In another study, \cite{wang2019SpeedRegressionUsing} proposed a deep CNN model for speed regression based on accelerometer data from wearable devices, achieving significant error reduction under different conditions. This study demonstrated the feasibility of deploying deep learning models on resource-constrained devices, reducing reliance on cloud-based processing. Moreover, advancements in neural network architectures have been pivotal. \cite{hannink2018MobileStrideLength} employed deep CNNs to predict stride length in elderly patients, showcasing the efficacy of deep learning in extracting meaningful features from inertial sensor data. Similarly, \cite{karlsson2021SpeedEstimationVibrations} introduced a CNN model using accelerometer and gyroscope data, outperforming traditional frequency analysis methods in automotive speed estimation tasks. These studies underline the superior performance of deep learning models in handling rapid speed changes and varying data rates, crucial for real-time applications.

Several works \citep{seethi2020CNNbasedSpeedDetection, bharti2019HuMAnComplexActivity} further refined CNN architectures for speed detection during walking and running, leveraging dual-signal-channel models combining accelerometer and gyroscope inputs. Their approach surpassed previous benchmarks, illustrating that integrating sensor inputs at different model layers enhances accuracy and reduces model size. This contrasts with earlier works like Wang's, which solely relied on accelerometer data, thereby highlighting the evolution towards more sophisticated sensor fusion techniques within ML frameworks.

While traditional sensor fusion methods remain foundational in speed estimation, the integration of ML techniques has revolutionised the field by offering robustness to complex environments and superior predictive capabilities. ML's adaptability to diverse sensor inputs and its ability to learn from large datasets without explicit modelling of dynamics are key advantages. However, challenges such as parameter tuning and computational resources remain, underscoring the ongoing need for optimising ML models for real-world deployment.

\section{Dataset preparation}
\label{sec:dataset_preparation}

Obtaining comprehensive datasets of train speeds presents significant challenges due to the commercially sensitive nature of such information. To develop an accurate algorithm for train status estimation, it is crucial to acquire not only precise train and wheel speeds but also GPS data and other sensor signals.

The datasets utilised in this article are extracted from the report ``Dependable Speed Measurement for Improved Low Adhesion Braking" \citep{brant2021DependableSpeedMeasurement} using image-to-data conversion software. The report implements 19 simulations, of which 17 are suitable for this study, as they have signals from both the wheel and the GPS. Specifically, 13 simulations are conducted without Wheel Slide Protection (WSP) operation, while 4 include the WSP operation. Two simulations are excluded due to the absence of GPS speed data.

\begin{figure}[htp!]
    \begin{subfigure}[h]{0.48\linewidth}
        \includegraphics[width=\linewidth]{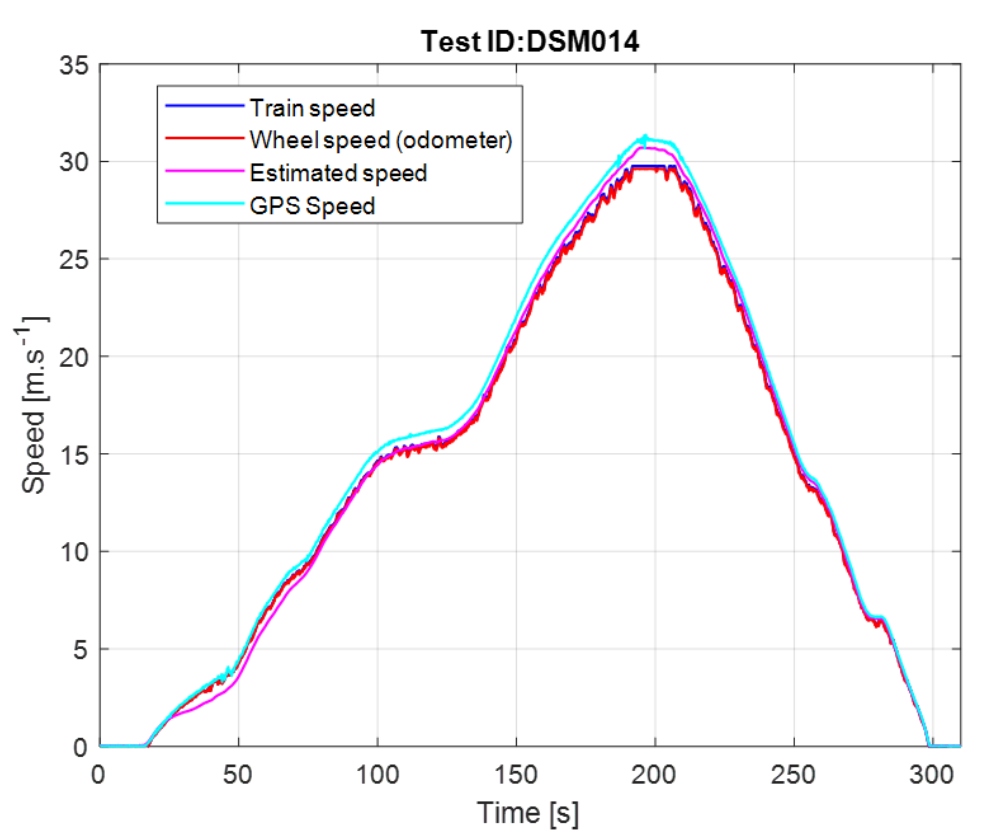}
        \caption{Report test DSM014. No odometer error, no WSP operation simulated.}
        \label{fig:dataset_example_1}
    \end{subfigure}%
    \hfill
    \begin{subfigure}[h]{0.48\linewidth}
        \includegraphics[width=\linewidth]{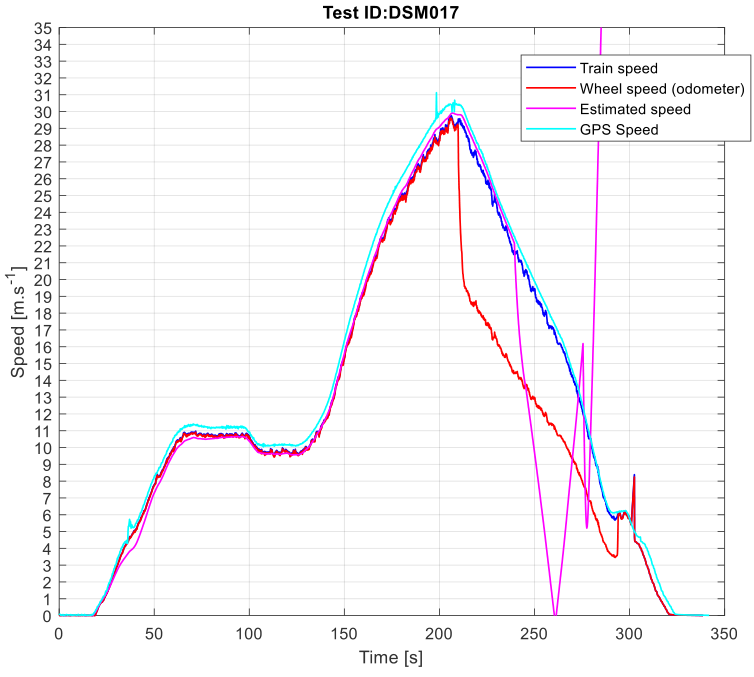}
        \caption{Report test DSM017. Odometer error introduced, no WSP operation simulated.}
        \label{fig:dataset_example_2}
    \end{subfigure}%
     
    \begin{subfigure}[h]{0.48\linewidth}
        \includegraphics[width=\linewidth]{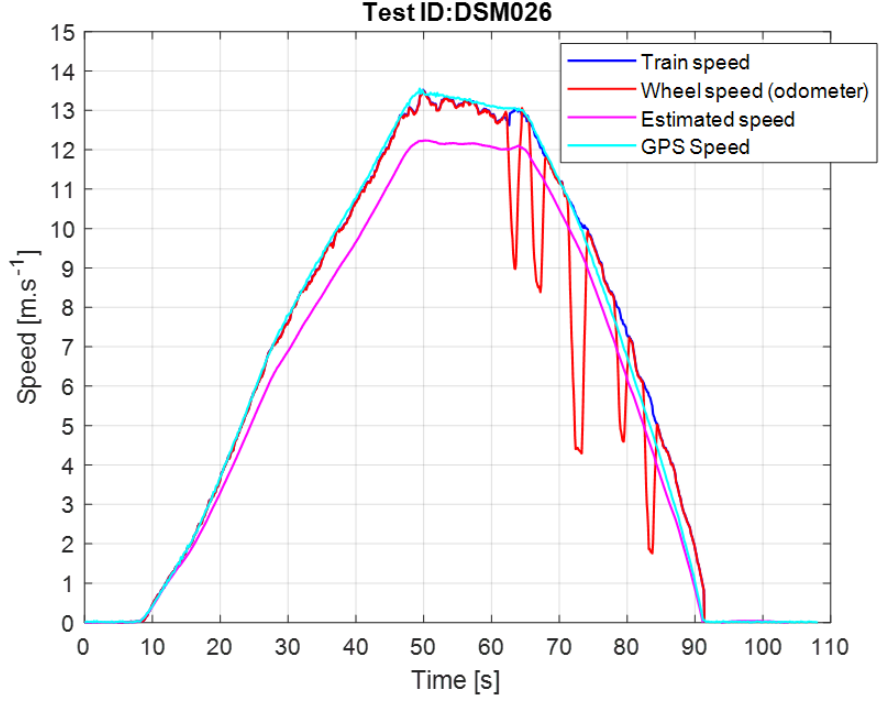}
        \caption{Report test DSM026. WSP operation simulated.}
        \label{fig:dataset_example_3}
    \end{subfigure}%
    \hfill
    \caption{Dependable Speed Measurement for Improved Low Adhesion Braking report figures}
    \label{fig:report}
\end{figure}

Figure \ref{fig:dataset_example_1} illustrates an example of the test figures presented in the report. The comparisons for each test are made between train speed, provided by the ground sensing radar; GPS speed, obtained from the INS input via GNSS antennas; wheel speed, inputted into the INS; and estimated speed, outputted from the INS. In this test, no odometer error is introduced and no WSP operation is simulated, resulting in the INS's estimated speed being quite accurate, with only minor errors during the acceleration phase and at the highest speed. Thus, the prediction task is relatively straightforward in this scenario.

Another example, shown in Figure \ref{fig:dataset_example_2}, demonstrates that during the acceleration phase, the estimated speed aligns with the train’s true speed. During the coasting phase, the estimated speed is slightly higher than the true train speed, with the GPS speed being approximately 1 m/s higher. At the beginning of the braking phase (190 seconds), a 20\% error is introduced to the odometer input of the INS (red line). At this point, the estimated speed matches the train’s true speed for 20 seconds before it starts to drift towards the odometer speed. This example indicates that the prediction task can potentially fail if the odometer reading contains significant error or noise.

Figure \ref{fig:dataset_example_3} presents a test example where the WSP operation is simulated. During the braking phase, the train wheels exhibit behaviour similar to the ABS system in cars, continuously braking and releasing, causing significant oscillations in the wheel speed readings. In such situations, the train driver cannot accurately determine the train’s operational state by reading the wheel speed, which can be very unsafe under certain circumstances. The figure also shows that the INS's speed estimation, which integrates wheel speed and GPS signals, displays some deviation. Accurately estimating train speed under these conditions becomes challenging, which is the primary issue addressed in this paper.

Four data channels are captured from the figures: time, wheel speed, GPS speed, and train speed, with the latter serving as the ground truth. Data from two figures are selected as test data - one without the WSP operation and one with the WSP operation. The remaining data are allocated for training and validation purposes. In total, the dataset comprises 5,205 time stamps for training and validation, and 947 for testing.

The algorithm's input is designed to incorporate not only the current speed information but also a specified number of historical speed data points. Assuming that the sampling frequency of the speed data is 1, then the input at a certain time step t goes from $S_{t-n+1}$ to $S_t$, where $S$ includes time, wheel speed, and GPS speed and $n$ is the history length used for each input. This approach enables the algorithm to learn from the evolving train status over time. Consequently, the size of the dataset may vary depending on the length of historical data considered. For instance, when using 10 as the history length, the training and validation set contains 5,055 data points. This number decreases to 4,905 when considering 20 as the history length, and further reduces to 4,755 for 30 history length. Despite these variations, the overall data sizes remain relatively consistent.

To standardise the data, wheel speeds and GPS speeds are normalised by dividing them by 31.3877 m/s, which represents the highest speed value in the datasets. In real-world scenarios or with different datasets, other thresholds can be used. 

The primary objective of this research is to identify an optimal CNN architecture that yields excellent prediction accuracy. The resulting model can be easily retrained to adapt to other datasets, ensuring its applicability across various scenarios.

\section{Methodology}
% Many methods have been developed to improve the accuracy of speed estimation. Conventional methods include the Kalman Filter, Particle Filter, etc. In contrast, machine learning-based speed estimation methods represent the cutting-edge techniques in this domain.

This section introduces the methodologies employed in this article for speed estimation. The first part examines the AKF, a well-established technique in conventional speed estimation. This method leverages recursive estimation and dynamic modelling to provide accurate train speed predictions, particularly in scenarios characterised by varying noise levels and changing dynamics. The second part explores the machine learning approach, specifically CNN. As a subset of deep learning, CNN offers a powerful alternative by learning complex patterns from large datasets and adapting to diverse sensor inputs.

\subsection{Conventional methods for speed estimation}
\label{sec:akf}

The Kalman Filter is a recursive estimation method, meaning that it can compute the current state estimate using only the estimate from the previous time step and the current observation \citep{govaers2019IntroductionImplementationsKalman}. Therefore, it does not need to retain a history of observations or estimates. Unlike most filters, the Kalman Filter is purely a time-domain filter. It does not require operating in the frequency domain followed by a conversion to the time domain, as is necessary for frequency-domain filters like low-pass filters.

The Kalman Filter model assumes that the true state at time $k$ evolves from the state at time $k-1$ according to the following equation:

\begin{equation}
    \mathbf{x}_k = \mathbf{F}_k \mathbf{x}_{k-1} + \mathbf{B}_k \mathbf{u}_k + \mathbf{w}_k
\end{equation}
where $\mathbf{F}_k$ is the state transition model applied to the previous state $\mathbf{x}_{k-1}$. $\mathbf{B}_k$ is the control-input model applied to the control vector $\mathbf{u}_k$. $\mathbf{w}_k$ is the process noise, which is assumed to follow a multivariate normal distribution with zero mean and covariance matrix $\mathbf{Q}_k$, denoted as $\mathbf{w}_k \sim N(0, \mathbf{Q}_k)$.

At time $k$, a measurement $\mathbf{z}_k$ of the true state $\mathbf{x}_k$ satisfies the following equation:

\begin{equation}
    \mathbf{z}_k = \mathbf{H}_k \mathbf{x}_k + \mathbf{v}_k
\end{equation}
where $\mathbf{H}_k$ is the observation model that maps the true state space into the observed space, and $\mathbf{v}_k$ is the observation noise, which is assumed to have zero mean and covariance matrix $\mathbf{R}_k$, denoted as $\mathbf{v}_k \sim N(0, \mathbf{R}_k)$.

The initial state and the noise terms at each time step $\{ \mathbf{x}_0, \mathbf{w}_1, \ldots, \mathbf{w}_k, \mathbf{v}_1, \ldots, \mathbf{v}_k \}$ are all assumed to be mutually independent.

Being a recursive estimator, the estimated current state is computed based only on the estimated state from the previous time step and the current measurement.

Two variables represent the state of the filter: $\hat{\mathbf{x}}_{k|k}$ (the a posteriori state estimate at time k given observations up to and including time k) and $\mathbf{P}_{k|k}$ (the a posteriori estimate covariance matrix, which measures the estimated accuracy of the state estimate).

The Adaptive Kalman Filter is an extension of the standard Kalman Filter designed to handle situations where the system model or noise characteristics are uncertain or time-varying. The AKF dynamically adjusts filter parameters based on observed data, allowing it to adapt to changing conditions.

The key difference in the AKF lies in how it estimates and updates the process noise covariance $\mathbf{Q}_k$ and measurement noise covariance $\mathbf{R}_k$. Two common approaches are covariance matching and maximum likelihood estimation. In the covariance matching approach, the innovation sequence is defined as:

\begin{equation}
\mathbf{\nu}_k = \mathbf{z}_k - \mathbf{H}_k \hat{\mathbf{x}}_{k|k-1}
\end{equation}
where $\hat{\mathbf{x}}_{k|k-1}$ is the a priori state estimate. The theoretical covariance of the innovation is:

\begin{equation}
\mathbf{S}_k = \mathbf{H}_k \mathbf{P}_{k|k-1} \mathbf{H}_k^T + \mathbf{R}_k
\end{equation}
where $\mathbf{P}_{k|k-1}$ is the a priori estimate covariance

The actual covariance of the innovation is estimated using a moving window of size N:

\begin{equation}
\hat{\mathbf{C}}_{\nu,k} = \frac{1}{N} \sum_{i=k-N+1}^k \mathbf{\nu}_i \mathbf{\nu}_i^T
\end{equation}

The noise covariances are then adapted:

\begin{equation}
\mathbf{R}_k = \hat{\mathbf{C}}_{\nu,k} - \mathbf{H}_k \mathbf{P}_{k|k-1} \mathbf{H}_k^T
\end{equation}

\begin{equation}
\mathbf{Q}_k = \mathbf{K}_k \hat{\mathbf{C}}_{\nu,k} \mathbf{K}_k^T
\end{equation}
where $\mathbf{K}_k$ is the Kalman gain.

In the maximum likelihood approach, the following log-likelihood function is maximised:

\begin{equation}
L = -\frac{1}{2} \sum_{k=1}^N \left( \ln |\mathbf{S}_k| + \mathbf{\nu}_k^T \mathbf{S}_k^{-1} \mathbf{\nu}_k \right)
\end{equation}

The resulting estimates of $\mathbf{Q}_k$ and $\mathbf{R}_k$ are then used in the standard Kalman Filter equations:

Prediction step:
\begin{equation}
\hat{\mathbf{x}}_{k|k-1} = \mathbf{F}_k \hat{\mathbf{x}}_{k-1|k-1} + \mathbf{B}_k \mathbf{u}_k
\end{equation}
\begin{equation}
\mathbf{P}_{k|k-1} = \mathbf{F}_k \mathbf{P}_{k-1|k-1} \mathbf{F}_k^T + \mathbf{Q}_k
\end{equation}

Update step:
\begin{equation}
\mathbf{K}_k = \mathbf{P}_{k|k-1} \mathbf{H}_k^T (\mathbf{H}_k \mathbf{P}_{k|k-1} \mathbf{H}_k^T + \mathbf{R}_k)^{-1}
\end{equation}
\begin{equation}
\hat{\mathbf{x}}_{k|k} = \hat{\mathbf{x}}_{k|k-1} + \mathbf{K}_k (\mathbf{z}_k - \mathbf{H}_k \hat{\mathbf{x}}_{k|k-1})
\end{equation}
\begin{equation}
\mathbf{P}_{k|k} = (\mathbf{I} - \mathbf{K}_k \mathbf{H}_k) \mathbf{P}_{k|k-1}
\end{equation}

The AKF iteratively updates these equations along with the noise covariance estimation, enabling it to adapt to changing system dynamics and noise characteristics. This approach allows the filter to maintain optimal performance even when faced with uncertainties or variations in the system model or noise properties.

\subsection{CNN-based method for speed estimation}

Convolutional Neural Networks \citep{atlas1987artificial,denker1988neural,lecun1989backpropagation,lecun1998gradient} are a class of neural networks characterized by the inclusion of convolutional operations \citep{ketkar2021ConvolutionalNeuralNetworks}.  Originally developed for image recognition, CNNs also show strong performance in time-series regression tasks by capturing local temporal patterns across multiple input channels.

In this study, CNNs are used to estimate train speed from multi-channel time-series windows. Given an input window $\mathbf{x} \in \mathbb{R}^{T \times C}$, where $T$ is the number of time steps and $C$ is the number of input channels (e.g., GPS speed, wheel speed), the CNN aims to learn a nonlinear mapping:

\begin{equation}
    v = f_\theta(\mathbf{x})
\end{equation}
where $f_\theta$ represents the CNN model with trainable parameters $\theta$, and $v$ is the estimated train speed.

The CNN architecture typically consists of stacked convolutional layers followed by nonlinear activation functions, pooling layers for dimensionality reduction, and fully connected layers for regression. This design allows the model to automatically extract and integrate spatial-temporal features of the input data. The core of the convolutional layer is a small matrix called a convolution kernel or filter. It slides over the input matrix and performs a simple weighted summation operation at each position to extract local features from the input data and produce the feature map. Assuming the input of the neural network is a 2D matrix $I$ and the convolution kernel is $K$, then the value at position $(i,j)$ in the output feature map after the convolution operation is:
\begin{equation}
    S(i,j) = (I * K)(i,j) = \sum_m \sum_n I(i+m, j+n) \cdot K(m, n)
\end{equation}

The activation function plays a key role in transforming linear inputs into nonlinear outputs, thereby enhancing the network's expressive power. Without activation functions, a multi-layer neural network would degrade into a simple linear model, incapable of capturing the complex nonlinear relationships within the data. This work employs the Rectified Linear Unit (ReLU) as the activation function, which is a piecewise linear function mathematically expressed as
\begin{equation}
    \text{ReLU}(x) = \max(0, x).
\end{equation}

The neural network is trained by minimising a mean squared error loss $L$:
\begin{equation}
    L=\frac{1}{N}\sum_{i=1}^{N}(y_i-\hat{y}_i)^2
\end{equation}
where $N$ is the number of samples in the mini-batch, $y_i$ is the ground truth, and $\hat{y}$ is the inferred output. The minimisation process is mini-batch stochastic gradient descent.

To improve generalisation and avoid overfitting, dropout layers are used during training. Hyperparameters such as the number of convolutional blocks, kernel size, learning rate, and dropout rate are optimised using the Optuna\footnote{\url{https://optuna.org/}} library, which applies Bayesian optimisation to efficiently explore the hyperparameter space. It employs the Tree-structured Parzen Estimator, a Bayesian optimisation-based algorithm, as its default hyperparameter tuning approach. Within this framework, kernel density estimation is utilised to model the probability density functions separately for the elite set, which comprises the better-performing hyperparameter configurations, and the non-elite set, consisting of the poorer-performing ones. This methodology allows Optuna to intelligently guide the search process towards regions of the hyperparameter space that are more likely to yield improved performance. 

Three CNN architectures inspired by the works of Karlsson \citep{karlsson2021SpeedEstimationVibrations}, Wang \citep{wang2019SpeedRegressionUsing}, and Seethi \citep{seethi2020CNNbasedSpeedDetection} are evaluated and adapted to our dataset, aiming to find the design that achieves the highest accuracy in speed estimation. Their specific details, together with the training details are presented in the following subsections.

\subsubsection{Single-branch 2D model.}
\label{sec:single_branch_2d_model}
The Single-branch 2D model proposed in \citep{karlsson2021SpeedEstimationVibrations} consists of three blocks, with each block containing a convolutional layer followed by a ReLU activation function and batch normalisation, succeeded by a 20\% dropout layer, and culminating in a fully connected layer and a regression layer. %The structure is illustrated in Figure \ref{fig:karlsson_original}. 
The input to the network is N × M, where M is the number of sensor signals used and N is the number of history time steps as explained in the Section titled \nameref{sec:dataset_preparation}.

A larger input shape allows for the inclusion of more historical information, while a smaller input shape can track rapid changes and reduce computational complexity. The input shapes examined for this architecture are 10x3, 20x3, 30x3, and 40x3, each incorporating three signals: time, wheel speed, and GPS speed.

Another critical parameter in the design of CNNs is the number of layers or blocks. A model with too few blocks may underfit the data, failing to capture the underlying features effectively. Conversely, a model with too many layers may overfit the training data, learning noise and specific details that do not generalise well to new data. Determining the optimal number of blocks involves balancing model complexity, performance, and computational efficiency. The original architecture employs three blocks, in which the batch normalisation layers are put after the activation functions ReLU. Studies indicate that ``adding the Batch Normalisation (BN) transform immediately before the nonlinearity" is ``likely to produce activations with a stable distribution"
\citep{ioffe2015BatchNormalizationAccelerating}. In this work, the BN layer is placed before the activation function ReLU, deviating from the original architecture. A range of 1 to 20 blocks is provided to Optuna to identify the optimal configuration.

To simplify the optimisation search problem, the number of filters and the kernel size in each convolutional layer are kept consistent across all blocks. The range of filters evaluated spans from 8 to 64. Additionally, three kernel sizes are tested: (3,2), (5,2), and (7,2).

Three additional hyperparameters are also evaluated: the dropout rate, ranging from 0.0 to 0.5; the learning rate, spanning from $10^{-5}$ to $ 10^{-2}$ on a logarithmic scale; and the batch size, with values of 8, 16, 32, and 64. These hyperparameters are considered to assess their impact on the prediction accuracy.

\subsubsection{Single-branch 1D model.}
\label{sec:single_branch_1d_model}

The second architecture considered in this work is one adapted by \cite{wang2019SpeedRegressionUsing} from a renowned deep CNN model originally developed for handwriting classification \citep{geron2023HandsonMachineLearning}. The 2-dimensional spatial convolution is replaced with 1-dimensional convolution to facilitate pattern recognition on time-series acceleration data, and the design parameters, including speed sample sizes, convolutional kernel sizes, and fully connected layer dimensions, are minimised to allow for a more compact hardware implementation.

The CNN architecture begins with an input tensor of shape (1,100,6), where 100 is the sequence length, and 6 is the number of input signals. The first convolutional layer processes this input and produces 32 feature maps. This is followed by a MaxPooling operation with a pool size of 1x5, which reduces the sequence length to 20 while retaining the same quantity of feature maps, resulting in a tensor of shape (1,20,32). A second convolutional layer is applied, maintaining the tensor shape at (1,20,32). Another MaxPooling operation with a pool size of 1x5 further reduces the sequence length to 4, yielding a tensor of shape (1,4,32). This output is then flattened and passed through a fully connected layer with 128 units, followed by a dropout layer. The subsequent fully connected layer with 32 units produces the final output tensor, which is then transformed into a single output speed value.

In the hyperparameter study for this architecture, four distinct input shapes are evaluated: (1,10,3), (1,20,3), (1,30,3), and (1,40,3). Each input shape is assessed to determine its impact on the performance. The number of blocks, where a block is defined as a convolutional layer followed by a MaxPooling layer, varies from 1 to 20. The kernel sizes considered range from 2 to 10. Other hyperparameters, including the number of filters, dropout rate, learning rate, and batch size, are maintained within the ranges previously established during the optimisation process for the previous model.

\subsubsection{Multiple-branch model.}
\label{sec:multiple_branch_model}

The final architecture evaluated and optimised is the Multiple-branch model \citep{seethi2020CNNbasedSpeedDetection}. %depicted in Figure \ref{fig:seethi_original}. 
A notable distinction of this structure from the previous two is that each sensor signal is processed independently through separate convolutional branches before the features are concatenated. This approach offers the advantage of mitigating noise introduced by signal fusion, thereby simplifying the feature extraction process. The concatenation operation integrates features passed by the global max-pooling layers from multiple branches, and outputs to two dense layers followed by a dropout layer respectively. Finally, a single neuron output layer presents the estimated speed.

Optuna searches the optimal input shape of each branch within four options: (10,1), (20,1), (30,1) and (40,1). A block in this architecture is defined as two consecutive convolutional layers, with the second layer having twice the number of filters as the first. This design ensures that each block's structure and the number of convolutional kernels remain similar to the original architecture. The number of blocks being searched ranges from 1 to 3, which is narrower than the range considered in the previous two architectures due to the presence of multiple channels. 

The structure following the convolutional layers remains consistent with the original architecture. The two dense layers contain 64 and 32 neurons, respectively, considering that the sample size in the reference is 156, whereas in this work, it is only 40 at most, which is significantly smaller. The ranges of number of filters, dropout rate and learning rate remain the same with the Single-branch 2D model and the Single-branch 1D model.

\section{Results}

\subsection{Adaptive Kalman Filter}
\begin{figure}[htp!]
    \begin{subfigure}[h]{0.48\linewidth}
        \includegraphics[width=\linewidth]{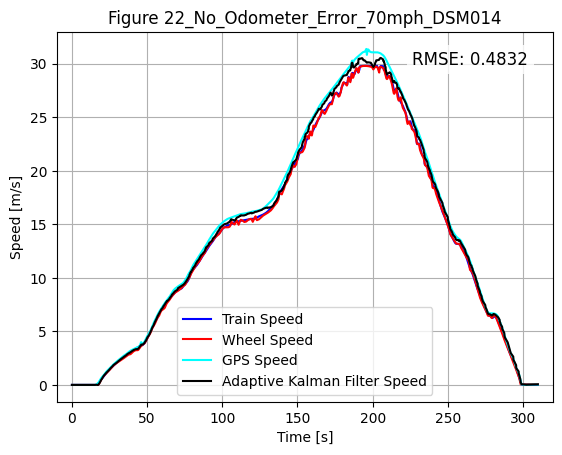}
        \caption{Train speed prediction of the test without WSP}
        \label{fig:prediction_akf_figure22}
    \end{subfigure}%
    \hfill
    \begin{subfigure}[h]{0.48\linewidth}
        \includegraphics[width=\linewidth]{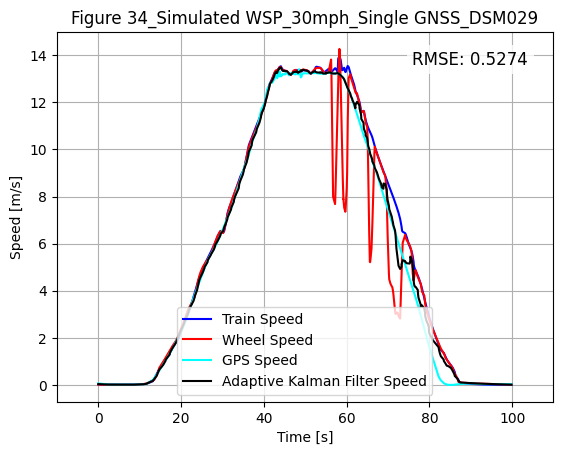}
        \caption{Train speed prediction of the test with WSP}
        \label{fig:prediction_akf_figure34}
    \end{subfigure}%
     
    \begin{subfigure}[h]{0.48\linewidth}
        \includegraphics[width=\linewidth]{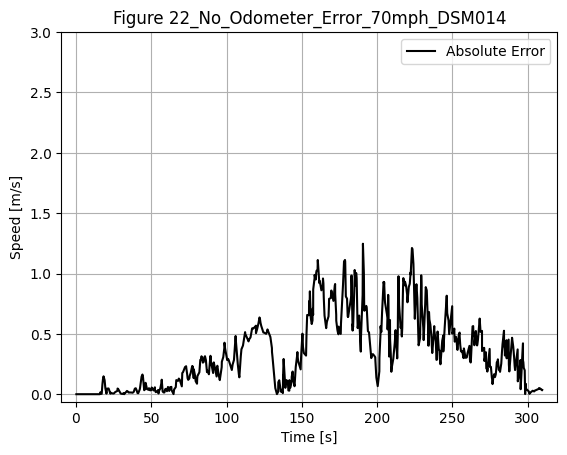}
        \caption{Prediction error of the test without WSP}
        \label{fig:prediction_akf_figure22_error}
    \end{subfigure}%
    \hfill
    \begin{subfigure}[h]{0.48\linewidth}
        \includegraphics[width=\linewidth]{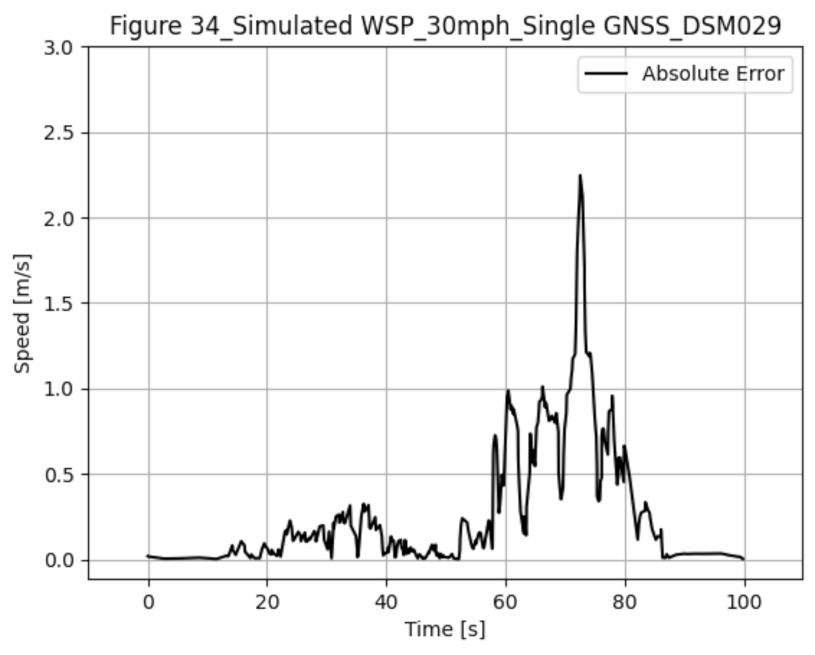}
        \caption{Prediction error of the test with WSP}
        \label{fig:prediction_akf_figure34_error}
    \end{subfigure}%
    \caption{Train speed predictions and errors by the AKF}
    \label{fig:prediction_akf}
\end{figure}

The AKF introduced in the section titled \nameref{sec:akf} is applied to predict train speeds on test DSM014 (without WSP operation), and test DSM029 (with WSP operation). The results in Figure \ref{fig:prediction_akf} provide a baseline for comparison with the subsequent results obtained from the CNN models.

The AKF algorithm demonstrates a good estimation of train speeds when the speed varies smoothly, particularly at lower speeds, as exemplified in the first 150s of Figure \ref{fig:prediction_akf_figure22} and the first 55s of Figure \ref{fig:prediction_akf_figure34}. However, Figure \ref{fig:prediction_akf_figure22_error} reveals that the error fluctuations become more pronounced when the train speed exceeds 20 m/s, occasionally exceeding 1m/s. The AKF algorithm encounters challenges when dealing with high train speeds.

After 55s in Figure \ref{fig:prediction_akf_figure34}, when the WSP operation is simulated, the AKF's prediction accuracy rapidly deteriorates, reaching a peak of 2.25m/s. The presence of WSP operation is associated with a higher Root Mean Square Error (RMSE) of 0.5274 compared to 0.4832 in its absence. The wheel slip scenario introduces nonlinear dynamics and uncertainties, which the AKF may not fully account for in its estimation process, even if the scene is in an overall low-speed state.

\subsection{CNN optimal hyperparameter search}

In this section are presented the results for the hyperparameter optimisation for the three CNN architectures studied. These results are required to define the final CNN architectures, which will lead to the machine learning results presented in the next section.

The hyperparameter search spaces are summarised in Table \ref{tab:hyperparameter_ranges}.

\begin{table}[ht]
    \small\sf\centering
    \caption{Hyperparameter ranges of CNN architectures}
    % \begin{tabular}{l|c|c|c}
    \begin{tabular}{|p{2.3cm}|p{1.5cm}|p{1.5cm}|p{1.5cm}|}
    \hline
      & Single-branch 2D & Single-branch 1D & Multiple-branch \\ \hline
    \multirow{2}{*}{Input shape} &(10,3,1), (20,3,1), & (10,3), (20,3),& (10,1), (20,1), \\
     & (30,3,1), (40,3,1) & (30,3),(40,3) &  (30,1), (40,1) \\
    Number of blocks & 1-20 & 1-20 & 1-3 \\
    Number of filters & 8-64 & 8-64 & 8-64 \\
    Kernel size & (3,2), (5,2), (7,2) & 2-10 & 2-10 \\
    Dropout rate & 0-0.5 & 0-0.5 & 0-0.5 \\
    Learning rate & $10^{-5}-10^{-2}$ & $10^{-5}-10^{-2}$ & $10^{-5}-10^{-2}$ \\
    Batch size & 8, 16, 32, 64 & 8, 16, 32, 64 & 8, 16, 32, 64 \\ \hline
    \end{tabular}
    \label{tab:hyperparameter_ranges}
\end{table}

For each architecture, a comprehensive study is conducted wherein a model is trained for 60 epochs in each trial, employing an 80/20 split ratio for the training and validation data. Throughout the training process, the performance of the model is monitored, and the best validation loss achieved during each trial is recorded.

\begin{figure}[htp!]
    \begin{subfigure}[h]{0.48\linewidth}
        \includegraphics[width=\linewidth]{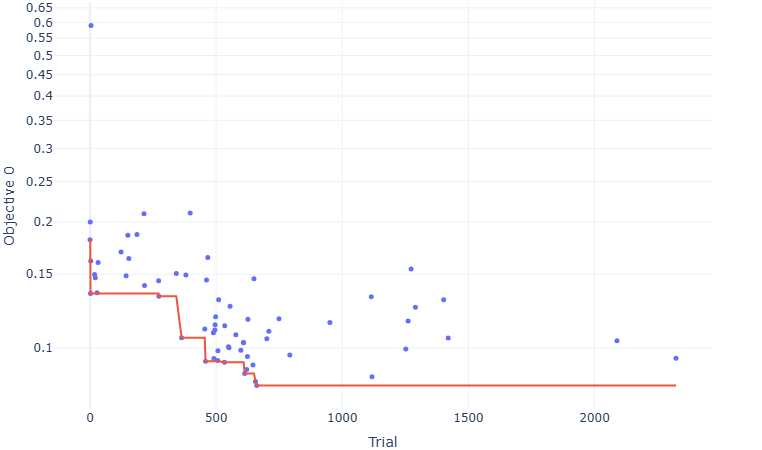}
        \caption{Study history for the Single-branch 2D model}
        \label{fig:optuna_for_karlsson_a}
    \end{subfigure}%
    \hfill
    \begin{subfigure}[h]{0.48\linewidth}
        \includegraphics[width=\linewidth]{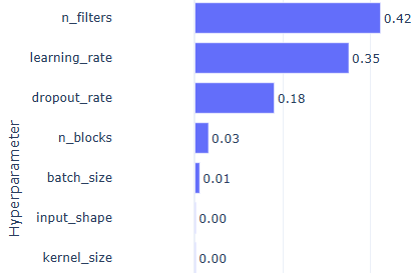}
        \caption{Hyperparameter importance for the Single-branch 2D model}
        \label{fig:optuna_for_karlsson_b}
    \end{subfigure}%

    \begin{subfigure}[h]{0.48\linewidth}
        \includegraphics[width=\linewidth]{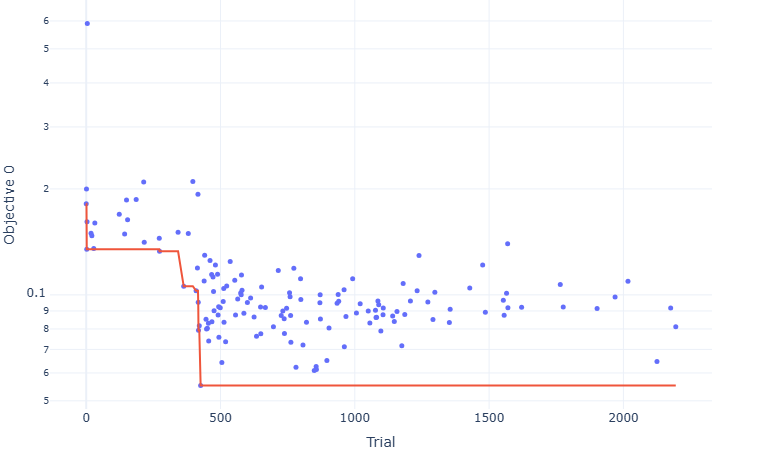}
        \caption{Study history for the Single-branch 1D model}
        \label{fig:optuna_for_wang_a}
    \end{subfigure}%
    \hfill
    \begin{subfigure}[h]{0.48\linewidth}
        \includegraphics[width=\linewidth]{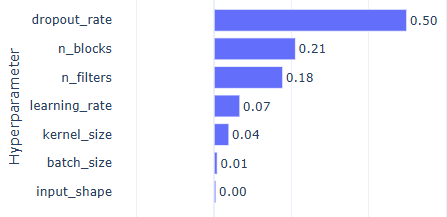}
        \caption{Hyperparameter importance for the Single-branch 1D model}
        \label{fig:optuna_for_wang_b}
    \end{subfigure}%
     
    \begin{subfigure}[h]{0.48\linewidth}
        \includegraphics[width=\linewidth]{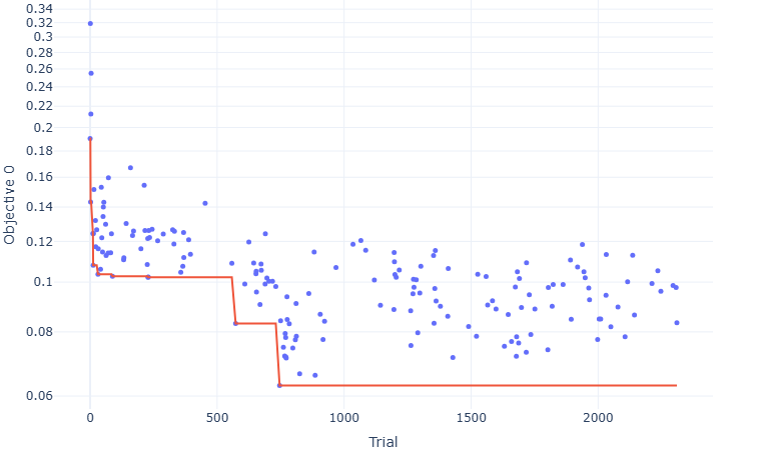}
        \caption{Study history for the Multiple-branch model}
        \label{fig:optuna_for_seethi_a}
    \end{subfigure}%
    \hfill
    \begin{subfigure}[h]{0.48\linewidth}
        \includegraphics[width=\linewidth]{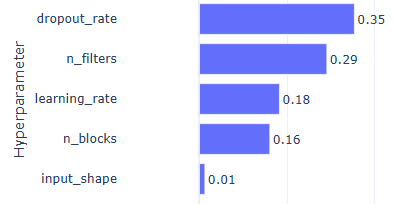}
        \caption{Hyperparameter importance for the Multiple-branch model}
        \label{fig:optuna_for_seethi_b}
    \end{subfigure}%
    \caption{Optimal hyperparameter search. In the left figures, the blue dots are completed trials and the red line represents the best validation loss at each point of the search. The right figures show the relative importance of each hyperparameter}
    \label{fig:optuna_three}
\end{figure}

The search history of optimal hyperparameters and their importance for the three architectures are plotted in Figure \ref{fig:optuna_three}. Pruned trials are excluded from the plot, and the y-axis is set to a logarithmic scale. In all three cases, we can observe a general downward trend in the objective value as the number of trials increases, indicating that the hyperparameter optimisation process is successfully finding better configurations over time. The red line in each graph represents the minimum validation loss value achieved so far, showing a step-wise improvement as better solutions are found.

The search histories of the three architectures exhibit similar patterns, characterised by significant drops in the objective value within the first 500 trials. The best values in all three cases stabilise after approximately 750 trials, with no further improvements observed thereafter. %The minimum validation losses achieved are 0.081, 0.055, and 0.063, respectively. However, a small validation loss value does not necessarily indicate superior predictive performance. This is because validation loss primarily measures the model's performance on a subset of the training data, and a lower validation loss can sometimes indicate overfitting, where the model becomes too tailored to the training data and fails to generalise well to unseen data.

In Figure \ref{fig:optuna_for_karlsson_a}, the number of filters (n\_filters) is the most important hyperparameter for the Single-branch 2D model, with an importance score of 0.42. This is closely followed by the learning rate at 0.35. The dropout rate also plays a significant role with a score of 0.18. The number of blocks (n\_blocks) and batch size have much lower importance scores, while input shape and kernel size appear to have negligible impact on the model's performance for this architecture. Figure \ref{fig:optuna_for_wang_a} presents a different distribution of importance. The dropout rate takes the lead as the most crucial hyperparameter with a score of 0.50. The number of blocks and the number of filters follow with moderate importance at 0.21 and 0.18 respectively. Other hyperparameters show a very small impact, suggesting minimal to no importance for the Single-branch 1D model. In Figure \ref{fig:optuna_for_seethi_a}, it can be seen that for the Multiple-branch model, the dropout rate and the number of filters are key factors, with the learning rate and the number of blocks contributing moderately.

Table \ref{tab:optimal_hyperparameters} summarises the optimal hyperparameters identified through the search within the defined spaces. In each case, over 2000 trials were conducted, with the majority of these trials being automatically pruned by the algorithm due to their large validation losses.

\begin{table}[ht]
    \caption{Optimal hyperparameters of three architectures searched}
    \begin{center}
        % \begin{tabular}{l|c|c|c}
        \begin{tabular}{|p{1.9cm}|p{1.5cm}|p{1.5cm}|p{1.5cm}|}
    \hline
         & Single-branch 2D & Single-branch 1D & Multiple-branch \\ \hline
        Total trials     & 2429 & 2196 & 2314 \\
        Pruned trials    & 2363 & 1994 & 2137 \\
        Completed trials &   66 &  202 &  177 \\
        Input shape & (20,3,1) & (10,3) & (30,1) \\
        Number of blocks & 3 & 4 & 2 \\
        Number of filters & 40 & 53 & 46 \\
        Kernel size & (7,2) & 2 & 2 \\
        Dropout rate & 4.9e-5 & 8.8e-3 & 1.9e-4 \\
        Learning rate & 1.7e-4 & 2.0e-3 & 1.8e-3 \\
        Batch size & 8 & 8 & 32 \\ \hline
        \end{tabular}
    \end{center}
    \label{tab:optimal_hyperparameters}
\end{table}

\subsection{Model training and results}
\label{sec:model_training_and_predictions}

In this section, the previous optimal hyperparameters are employed to train the three architectures. Subsequently, the models are applied to the test data to evaluate the predictive performance of each architecture.

\subsubsection{Single-branch 2D model.}

\begin{figure}[htp!]
    \centering
    \includegraphics[width=1.0\linewidth]{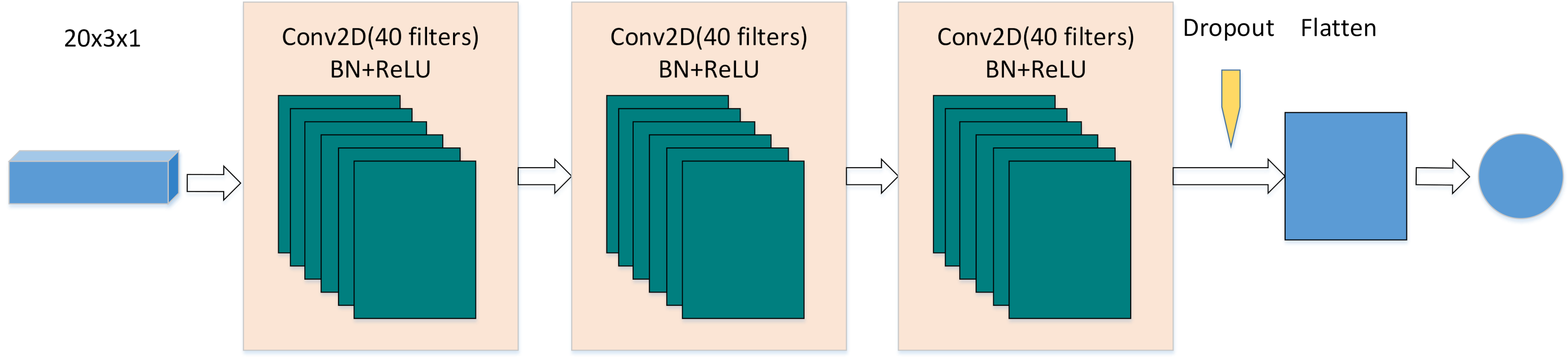}
    \caption{Structure of the Single-branch 2D model}
    \label{fig:karlsson_no_kfold_optimal_structure}
\end{figure}

\begin{figure}[htp!]
    \centering
    \includegraphics[width=1.0\linewidth]{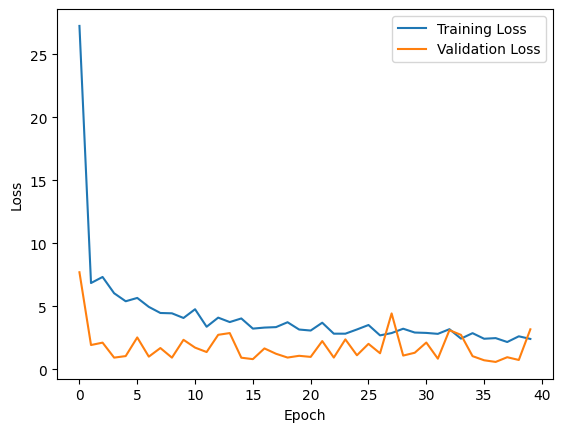}
    \caption{Training and validation loss of the Single-branch 2D model with optimal hyperparameters}
    \label{fig:karlsson_no_kfold_optimal_convergence}
\end{figure}

The Single-branch 2D model is trained with the hyperparameters listed in Table \ref{tab:optimal_hyperparameters}, and the structure plotted in Figure \ref{fig:karlsson_no_kfold_optimal_structure}. The model is trained for 40 epochs. The training process, as depicted in Figure \ref{fig:karlsson_no_kfold_optimal_convergence}, shows a downward trend as the model learns. The training loss curve maintains a relatively steady decline, which means the model is continuously improving its predictions on the training data. The validation loss curve, while also showing an overall decrease, experiences spikes at certain points because of a non-smooth loss landscape. %These spikes indicate that the model might be overfitting on the training data, meaning it is learning patterns that are too specific to the training data and are not generalisable to unseen data.

\begin{figure}[htp!]
    \begin{subfigure}[h]{0.48\linewidth}
        \includegraphics[width=\linewidth]{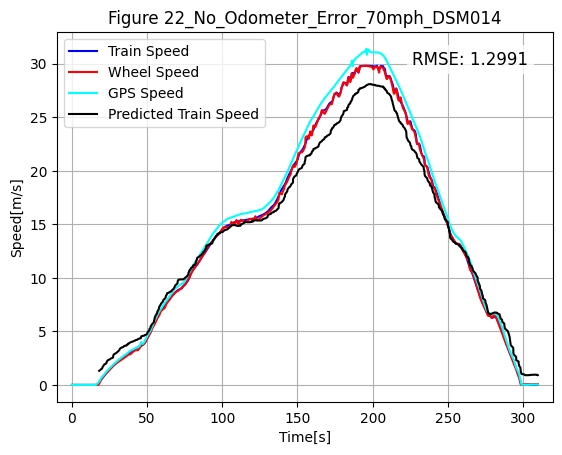}
        \caption{Train speed prediction of the test without WSP}
        \label{fig:prediction_karlsson_figure22}
    \end{subfigure}%
    \hfill
    \begin{subfigure}[h]{0.48\linewidth}
        \includegraphics[width=\linewidth]{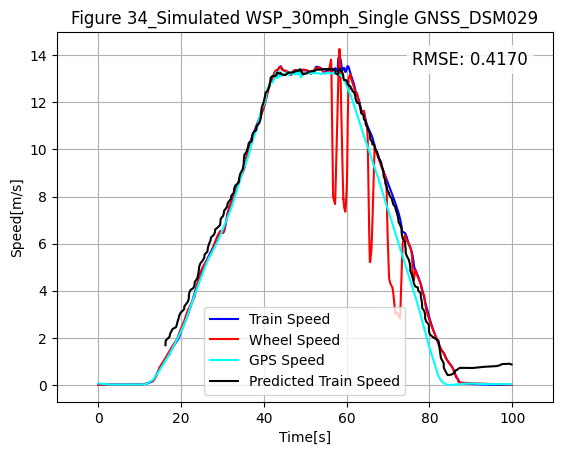}
        \caption{Train speed prediction of the test with WSP}
        \label{fig:prediction_karlsson_figure34}
    \end{subfigure}%
    
    \begin{subfigure}[h]{0.48\linewidth}
        \includegraphics[width=\linewidth]{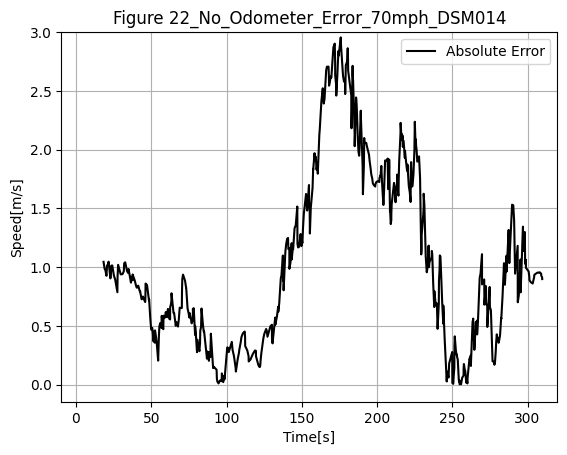}
        \caption{Prediction error of the test without WSP}
        \label{fig:prediction_karlsson_figure22_error}
    \end{subfigure}%
    \hfill
    \begin{subfigure}[h]{0.48\linewidth}
        \includegraphics[width=\linewidth]{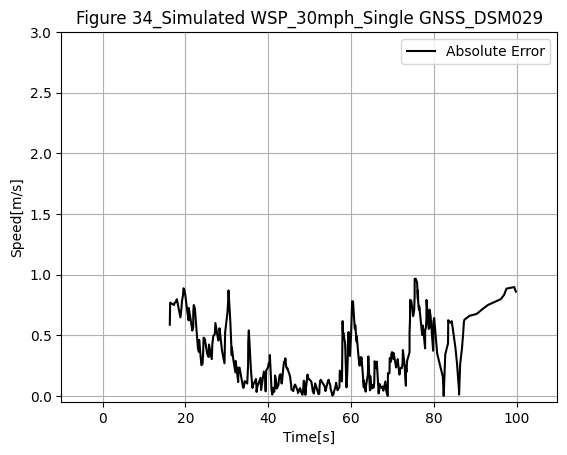}
        \caption{Prediction error of the test with WSP}
        \label{fig:prediction_karlsson_figure34_error}
    \end{subfigure}%
    \caption{Train speed predictions and errors by the Single-branch 2D model}
    \label{fig:prediction_karlsson}
\end{figure}

 Figure \ref{fig:prediction_karlsson} illustrates the true train speed, GPS speed, wheel speed, and the estimated train speed by the Single-branch 2D model for test cases DSM014 and DSM029, with a braking curve from 31 m/s and 13.5 m/s respectively. The model does not provide speed estimates in the first few seconds as it requires 20 consecutive time steps of speed data to generate predictions. In Figure \ref{fig:prediction_karlsson_figure22} and Figure \ref{fig:prediction_karlsson_figure22_error} for the test DSM014, where the wheel speed aligns closely with the train speed, the predicted speed is slightly higher than the ground truth until the train accelerates to 15 m/s, with an error not exceeding 1 m/s. However, when the train speed ranges from 15 m/s to 31 m/s, the prediction error reaches up to 3 m/s during both the acceleration and braking phases. The error reduces to less than 1.5 m/s in the final stage, but a residual error remains even when the train is at a complete stop. The prediction has a root mean square error of around 1.3 across the entire profile.

The speed profile of the test DSM029 in Figure \ref{fig:prediction_karlsson_figure34}, which includes the simulated WSP operation, is supposed to be more challenging to predict due to the significant discrepancies introduced by the wheel speed signal. Surprisingly, the network produces a better estimation with an error of no more than 1 m/s and a RMSE of 0.4170, even during the braking phase when the wheel speed deviates substantially from the ground truth. Similar to the test DSM014, the network ultimately predicts a non-zero speed when the train comes to a stop.

\subsubsection{Single-branch 1D model.}

\begin{figure}[htp!]
    \centering
    \includegraphics[width=1.0\linewidth]{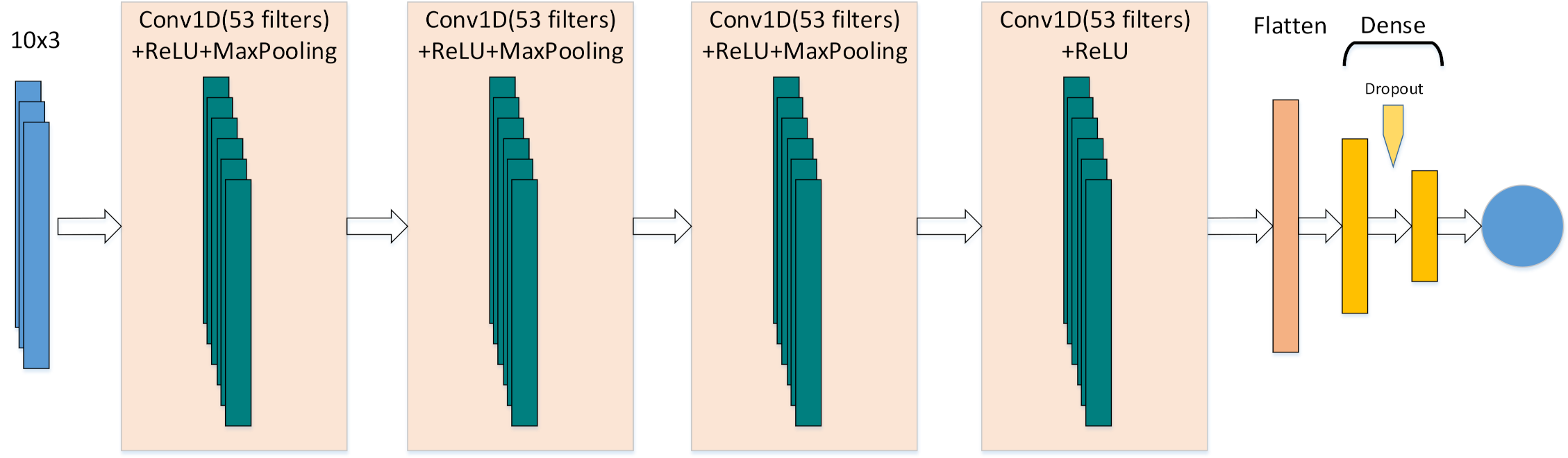}
    \caption{Structure of the Single-branch 1D model}
    \label{fig:wang_optimal_structure}
\end{figure}

\begin{figure}[htp!]
    \begin{center}
        \includegraphics[width=1.0\linewidth]{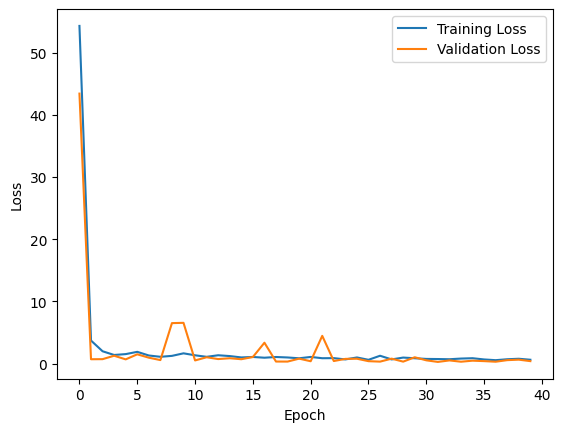}
    \end{center}
    \caption{Training and validation loss of the Single-branch 1D model with optimal hyperparameters}
    \label{fig:wang_no_kfold_optimal_convergence}
\end{figure}

By training the optimal hyperparameters and structure of the Single-branch 1D model plotted in Figure \ref{fig:wang_optimal_structure}, the training and validation losses illustrated in Figure \ref{fig:wang_no_kfold_optimal_convergence} are obtained. It is clear that this model converges rapidly and exhibits significantly greater stability compared to the Single-branch 2D model. The loss curves also suggest that 40 epochs are sufficient, as additional training epochs do not further reduce the validation loss.

\begin{figure}[htp!]
    \begin{subfigure}[h]{0.48\linewidth}
        \includegraphics[width=\linewidth]{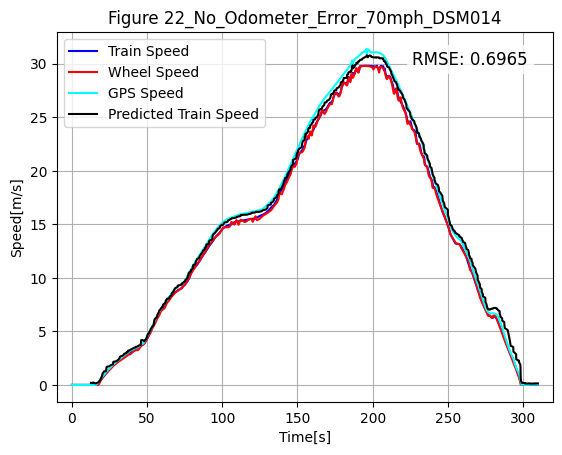}
        \caption{Train speed prediction of the test without WSP}
        \label{fig:prediction_wang_figure22}
    \end{subfigure}%
    \hfill
    \begin{subfigure}[h]{0.48\linewidth}
        \includegraphics[width=\linewidth]{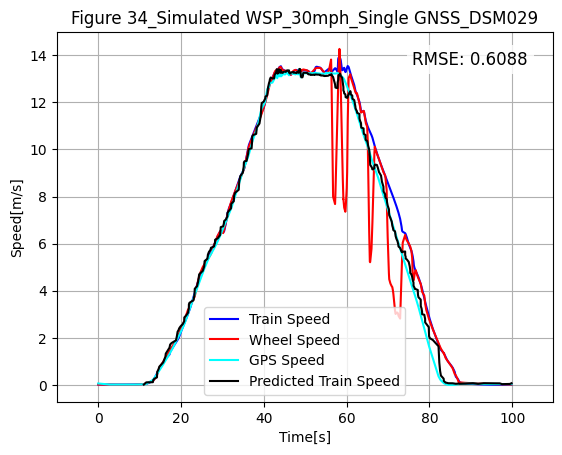}
        \caption{Train speed prediction of the test with WSP}
        \label{fig:prediction_wang_figure34}
    \end{subfigure}%
    
    \begin{subfigure}[h]{0.48\linewidth}
        \includegraphics[width=\linewidth]{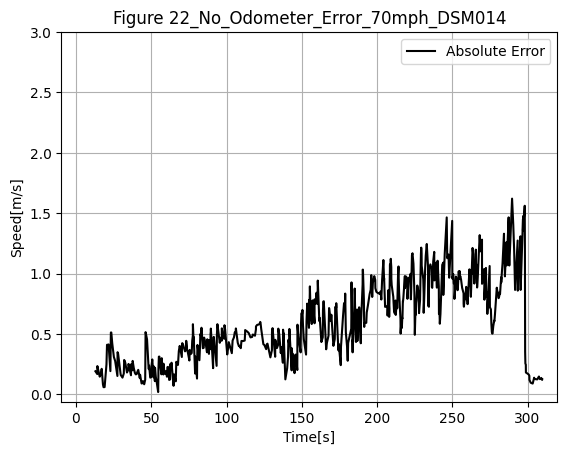}
        \caption{Prediction error of the test without WSP}
        \label{fig:prediction_wang_figure22_error}
    \end{subfigure}%
    \hfill
    \begin{subfigure}[h]{0.48\linewidth}
        \includegraphics[width=\linewidth]{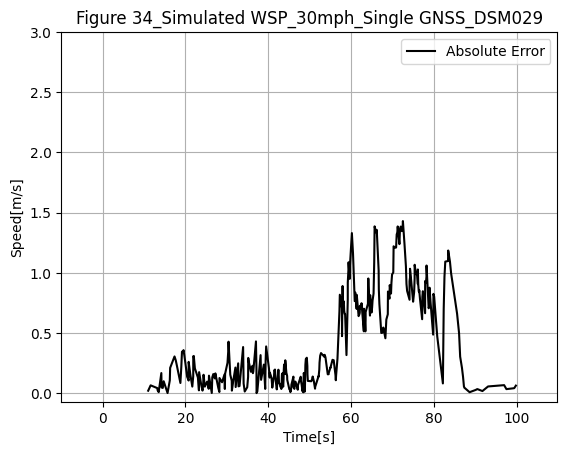}
        \caption{Prediction error of the test with WSP}
        \label{fig:prediction_wang_figure34_error}
    \end{subfigure}%
    \caption{Train speed predictions and errors by the Single-branch 1D model}
    \label{fig:prediction_wang}
\end{figure}

Figure \ref{fig:prediction_wang} illustrates the model's predictions for two tests. Compared to the previous Single-branch 2D model, this model performs quite well on DSM014 without the WSP operation, closely matching the true train speed throughout the entire profile with a maximum error of around 1.5 m/s and errors less than 1 m/s during the acceleration phase. The RMSE is reduced from 1.2991 to 0.6965. For DSM029 with the WSP operation, the errors are all below 1.5 m/s, with errors during the acceleration and constant speed phases smaller than 0.5 m/s. Notably, the model accurately predicts the zero-speed state after the train stops, which is more aligned with real-world conditions. Although the RMSE of this test is higher than that of the Single-branch 2D model and the prediction errors increase during the braking phase, the model demonstrates excellent performance during the acceleration and constant speed phases.

\subsubsection{Multiple-branch model.}

\begin{figure}[htp!]
    \centering
    \includegraphics[width=1.0\linewidth]{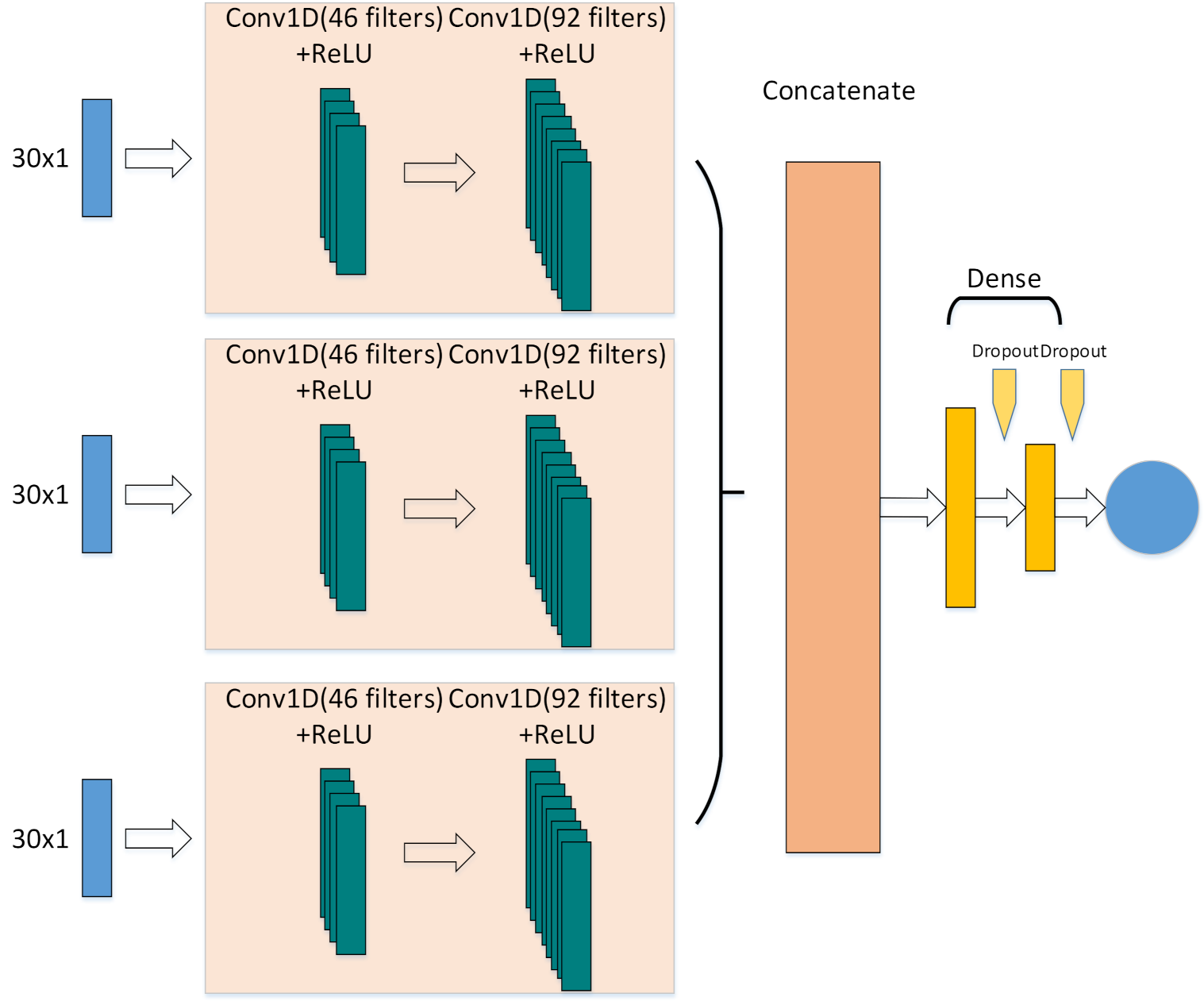}
    \caption{Structure of the Multiple-branch model}
    \label{fig:seethi_optimal_structure}
\end{figure}

The optimal structure of the Multiple-branch model is plotted in Figure \ref{fig:seethi_optimal_structure}.

\begin{figure}[htp!]
    \begin{center}
        \includegraphics[width=1.0\linewidth]{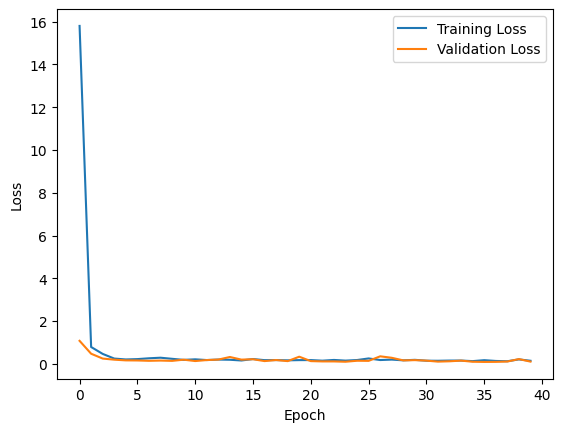}
    \end{center}
    \caption{Training and validation loss of the Multiple-branch model with optimal hyperparameters}
    \label{fig:seethi_no_kfold_optimal_convergence}
\end{figure}

 Figure \ref{fig:seethi_no_kfold_optimal_convergence} presents the losses of the model during its training. The flat curves indicate that the model has achieved a good fit on the training data and that its performance on the validation set is stable. It significantly outperforms the Single-branch 2D model, and the validation loss exhibits fewer anomalies compared to the Single-branch 1D model. Similar to the previous model, 40 epochs of training are sufficient to obtain stable results.

\begin{figure}[htp!]
    \begin{subfigure}[h]{0.48\linewidth}
        \includegraphics[width=\linewidth]{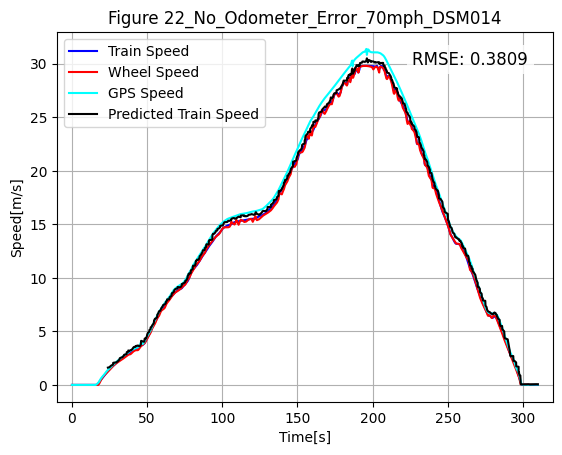}
        \caption{Train speed prediction of the test without WSP}
        \label{fig:prediction_seethi_figure22}
    \end{subfigure}%
    \hfill
    \begin{subfigure}[h]{0.48\linewidth}
        \includegraphics[width=\linewidth]{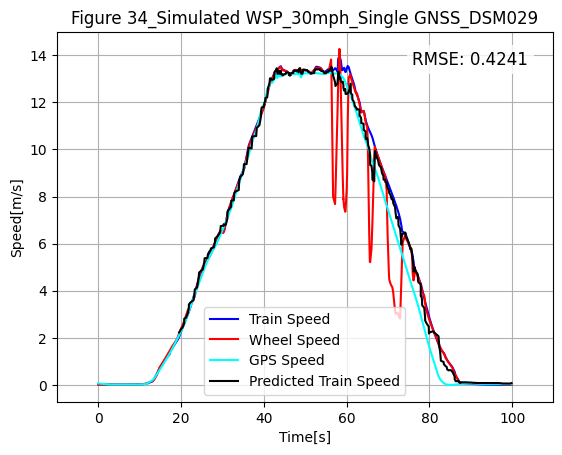}
        \caption{Train speed prediction of the test with WSP}
        \label{fig:prediction_seethi_figure34}
    \end{subfigure}%
    
    \begin{subfigure}[h]{0.48\linewidth}
        \includegraphics[width=\linewidth]{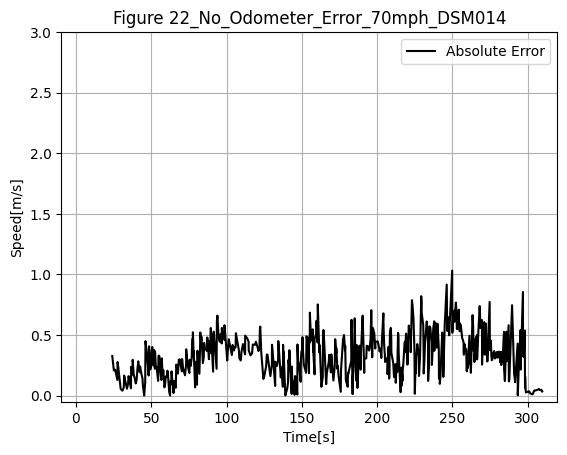}
        \caption{Prediction error of the test without WSP}
        \label{fig:prediction_seethi_figure22_error}
    \end{subfigure}%
    \hfill
    \begin{subfigure}[h]{0.48\linewidth}
        \includegraphics[width=\linewidth]{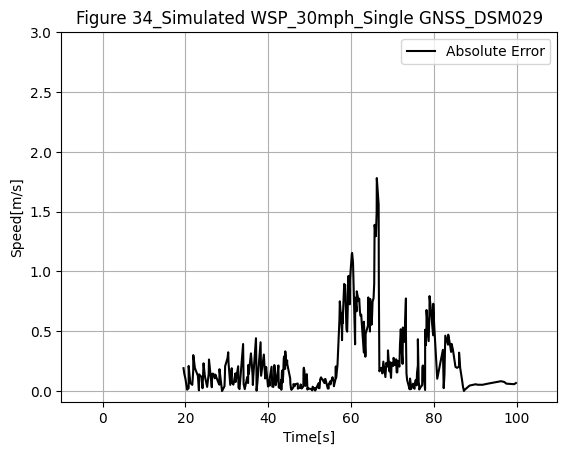}
        \caption{Prediction error of the test with WSP}
        \label{fig:prediction_seethi_figure34_error}
    \end{subfigure}%
    \caption{Train speed predictions and errors by the Multiple-branch model}
    \label{fig:prediction_seethi}
\end{figure}

 Figure \ref{fig:prediction_seethi} illustrates this model’s predictions on tests without and with the WSP operation. The RMSE for both tests is notably low. As shown in Figure \ref{fig:prediction_seethi_figure22} and Figure \ref{fig:prediction_seethi_figure22_error}, the estimated speed closely matches the train’s true speed throughout the entire test, with errors not exceeding 1 m/s in both the acceleration and braking phases. The model’s prediction accuracy does not significantly decrease at high speeds. In Figure \ref{fig:prediction_seethi_figure34}, before the introduction of the WSP operation, the prediction error remains below 0.5 m/s. The rapid changes in wheel speed increase the error slightly, but it generally stays below 1 m/s, with only one peak exceeding 1.5 m/s. The model's predictions for the train speed in both tests have very low RMSE values, respectively 0.3809 and 0.4241. Overall, the model based on the multiple-branch structure demonstrates high prediction accuracy and stability. This can be attributed to its independent processing of signals from different channels, which allows for the aggregation of all signal features and mitigates the impact of a single channel's failure on the final prediction outcome.

\subsection{Discussion}

The results presented in the Section titled \nameref{sec:model_training_and_predictions} indicate that the Single-branch 2D model demonstrates a relatively bad prediction performance. Its predictions exhibit substantial errors and high variability, particularly under smooth, low-speed train conditions. Although this model shows a lower RMSE with the WSP operation simulated, the reliability and stability of this architecture require further enhancement. Processing multiple signals as an image does not seem to be an effective approach. In contrast, the Single-branch 1D model provides more stable predictions during smooth acceleration and deceleration phases. However, it exhibits higher errors during the WSP operation simulation. Compared to the previous two models, the Multiple-branch model, which trains three separate signals separately, achieves the lowest prediction error under the scenario without WSP. When WSP operation simulated, its prediction accuracy and stability are slightly inferior to the Single-branch 2D model, but still comparable. The Multiple-branch model demonstrates the most robust performance among the three architectures.

The conventional speed estimation method - AKF - exhibits commendable prediction accuracy and stability, surpassing both the Single-branch 2D and the Single-branch 1D models. However, the CNN-based Multiple-branch model demonstrates greater potential. It achieves a smaller RMSE and lower maximum errors compared to the AKF. Further improvements in prediction accuracy and reductions in error peaks can be achieved by incorporating additional signals and expanding the dataset.

It is worth to note that several limitations that could impact the generalisability and robustness of the results. The work is based on simulated data. The dataset used is extracted from figures in an existing report, which introduces certain inaccuracies and limitations in the data. The size of the dataset is also constrained, which likely influenced the training efficacy and predictive performance of the CNN models. The current study focuses on only three channels of signal data. While these channels provide valuable information, the exclusion of other potentially relevant channels, such as acceleration and braking status, may have limited the model's robustness and reliability. Incorporating additional channels can enrich the dataset, allowing the CNN models to capture a broader range of operational dynamics. In addition, exploring the application of the model in certain channel signal loss scenarios can also improve the robustness of the system. These factors underscore the need for more accurate and comprehensive datasets in order to optimise the models further and achieve better prediction accuracy.

\section{Conclusions}

This study demonstrates the potential of Convolutional Neural Networks in the prediction of train speeds using signal data, offering a promising approach for advancing train control systems. The CNN models developed in this work were able to identify and learn patterns from the provided signal data, highlighting their applicability in the field of railway operations. Among the three CNN designs, the Multiple-branch model exhibits excellent performance in the scenario without WSP and maintains good predictive accuracy with WSP operation simulated. Overall, this model outperforms the other two CNN architectures and the Adaptive Kalman Filter in terms of robustness and accuracy across different train operating conditions.

The study's findings have important implications for railway safety and efficiency. By providing more accurate and robust speed estimates, especially in adverse conditions, these CNN-based models can enhance the performance of critical systems such as Automatic Train Protection and improve overall train control. 

Future work can focus on validating these models with real-world operational data across a wider range of conditions and train types. Further explorations may also include the integration of these CNN models into existing train control systems and the investigation of their real-time performance and computational requirements.

\section{Acknowledgments}
The work stems from an MSc Applied AI programme individual research project outcome supported by RSSB. The authors thank M. Carmichael for his invaluable guidance and support throughout the MSc project. The authors also thank the project reviewers for the useful discussions and comments during the work. 

\section{Funding}
The authors received no financial support for the research, authorship, and/or publication of this article.

%% Back matter
%
% This is where we include references and appendices

%%Harvard (name/date)
% \bibliographystyle{SageH}
%%Vancouver (numbered)
\bibliographystyle{SageV}

\end{document}